\documentclass[10pt, a4paper]{article}

\usepackage {lrec-coling2024}
\usepackage{times}
\usepackage{latexsym}
\usepackage{graphicx}
\usepackage{amsmath}
\usepackage{multirow}
\usepackage{subfigure}
\usepackage{bbding}
\usepackage{booktabs}
\usepackage[normalem]{ulem}
\useunder{\uline}{\ul}{}
\usepackage{bm}
\usepackage{inconsolata}
\usepackage[ruled,linesnumbered]{algorithm2e}
\usepackage{hyperref}
\newcommand{\rmnum}[1]{\romannumeral #1}

\usepackage{color}
\usepackage{fdsymbol}
\newcommand{\Thanks}[1]{\thanks{\ #1}}
\AtBeginDocument{
\def\maketitle{\par
 \begingroup
   \def\thefootnote{\fnsymbol{footnote}}
   \def\@makefnmark{\hbox to 0pt{$^{\@thefnmark}$\hss}}
   \twocolumn[\@maketitle] \@thanks
 \endgroup
 \setcounter{footnote}{0}
 \let\maketitle\relax \let\@maketitle\relax
 \gdef\@thanks{}\gdef\@author{}\gdef\@title{}\let\thanks\relax}
\def\@maketitle{\vbox to \titlebox{\hsize\textwidth
 \linewidth\hsize \vskip 0.125in minus 0.125in \centering
 {\Large\bf \@title \par} \vskip 0.2in plus 1fil minus 0.1in
 {\def\and{\unskip\enspace{\rm and}\enspace}%
  \def\And{\end{tabular}\hss \egroup \hskip 1in plus 2fil 
           \hbox to 0pt\bgroup\hss \begin{tabular}[t]{c}\bf}%
  \def\AND{\end{tabular}\hss\egroup \hfil\hfil\egroup
          \vskip 0.25in plus 1fil minus 0.125in
           \hbox to \linewidth\bgroup\large \hfil\hfil
             \hbox to 0pt\bgroup\hss \begin{tabular}[t]{c}\bf}
  \hbox to \linewidth\bgroup\large \hfil\hfil
    \hbox to 0pt\bgroup\hss 
	\outauthor
   \hss\egroup
    \hfil\hfil\egroup}
  \vskip 0.3in plus 2fil minus 0.1in
}}
}

\title{Unleashing the Power of Imbalanced Modality Information for Multi-modal Knowledge Graph Completion}

\name{Yichi Zhang$^{\clubsuit\heartsuit}$, Zhuo Chen$^{\clubsuit\heartsuit}$, Lei Liang$^{\diamondsuit}$, Huajun Chen$^{\clubsuit\heartsuit\spadesuit}$, Wen Zhang$^{\clubsuit\heartsuit\dag}$ \Thanks{\dag~~Corresponding author.}}
\address{$^\clubsuit$Zhejiang University, $^{\diamondsuit}$Ant Group\\
$^\spadesuit$Alibaba-Zhejiang University Joint Institute of Frontier Technology\\
$^\heartsuit$Zhejiang University-Ant Group Joint Laboratory of Knowledge Graph\\
        \{zhangyichi2022, zhuo.chen, zhang.wen\}@zju.edu.cn\\}

\abstract{
Multi-modal knowledge graph completion (MMKGC) aims to predict the missing triples in the multi-modal knowledge graphs by incorporating structural, visual, and textual information of entities into the discriminant models. The information from different modalities will work together to measure the triple plausibility. Existing MMKGC methods overlook the imbalance problem of modality information among entities, resulting in inadequate modal fusion and inefficient utilization of the raw modality information. To address the mentioned problems, we propose \textbf{Ada}ptive \textbf{M}ulti-modal \textbf{F}usion and \textbf{M}odality \textbf{A}dversarial
\textbf{T}raining (AdaMF-MAT) to unleash the power of imbalanced modality information for MMKGC. AdaMF-MAT achieves multi-modal fusion with adaptive modality weights and further generates adversarial samples by modality-adversarial training to enhance the imbalanced modality information. Our approach is a co-design of the MMKGC model and training strategy which can outperform 19 recent MMKGC methods and achieve new state-of-the-art results on three public MMKGC benchmarks. Our code and data have been released at \href{https://github.com/zjukg/AdaMF-MAT}{https://github.com/zjukg/AdaMF-MAT}.
\\ \newline  }

\begin{document}

\maketitleabstract

\section{Introduction}
Knowledge graphs (KGs) \cite{DBLP:journals/tkde/Survey} model the world knowledge as structured triples in the form of (\textit{head entity, relation, tail entity}). Multi-modal knowledge graphs (MMKGs) \citetlanguageresource{DBLP:conf/esws/MMKG} further extend KGs with representative multi-modal information (e.g. textual descriptions, images, etc.) and have become the new infrastructure in many artificial intelligence tasks \cite{DBLP:conf/cikm/MMKGRec,DBLP:conf/naacl/QAGNN,DBLP:conf/mm/MMKGusage3,DBLP:journals/corr/abs-2311-06503-knowpat,liang2023learn}.

However, KGs often suffer from incompleteness as there are many unobserved triples in the KGs. The incompleteness of KGs limits their usage in the downstream tasks. Therefore, knowledge graph completion (KGC) \cite{DBLP:journals/tkde/Survey} has attracted wide attention as a crucial task to complete the missing triples in the KGs. Mainstream KGC methods employ knowledge graph embedding (KGE) to model the structural information in the KGs, which 
embeds the entities and relations into a low-dimensional continuous space. 

\par As for the MMKGs, the KGC models \cite{DBLP:conf/mm/RSME,DBLP:conf/emnlp/MOSE} incorporate rich-semantic multi-modal information such as images and text descriptions, referred to as multi-modal knowledge graph completion (MMKGC). Mainstream MMKGC methods \cite{DBLP:conf/mm/RSME} usually focus on designing elegant methods to achieve multi-modal fusion among structural, visual, and textual information, which is a common way of thinking for multi-modal tasks. Some other methods \cite{DBLP:conf/mm/MMRNS} attempt to enhance the negative sampling (NS) process with the multi-modal information. 
\begin{figure}[]
    \centering
    \includegraphics[width=\linewidth]{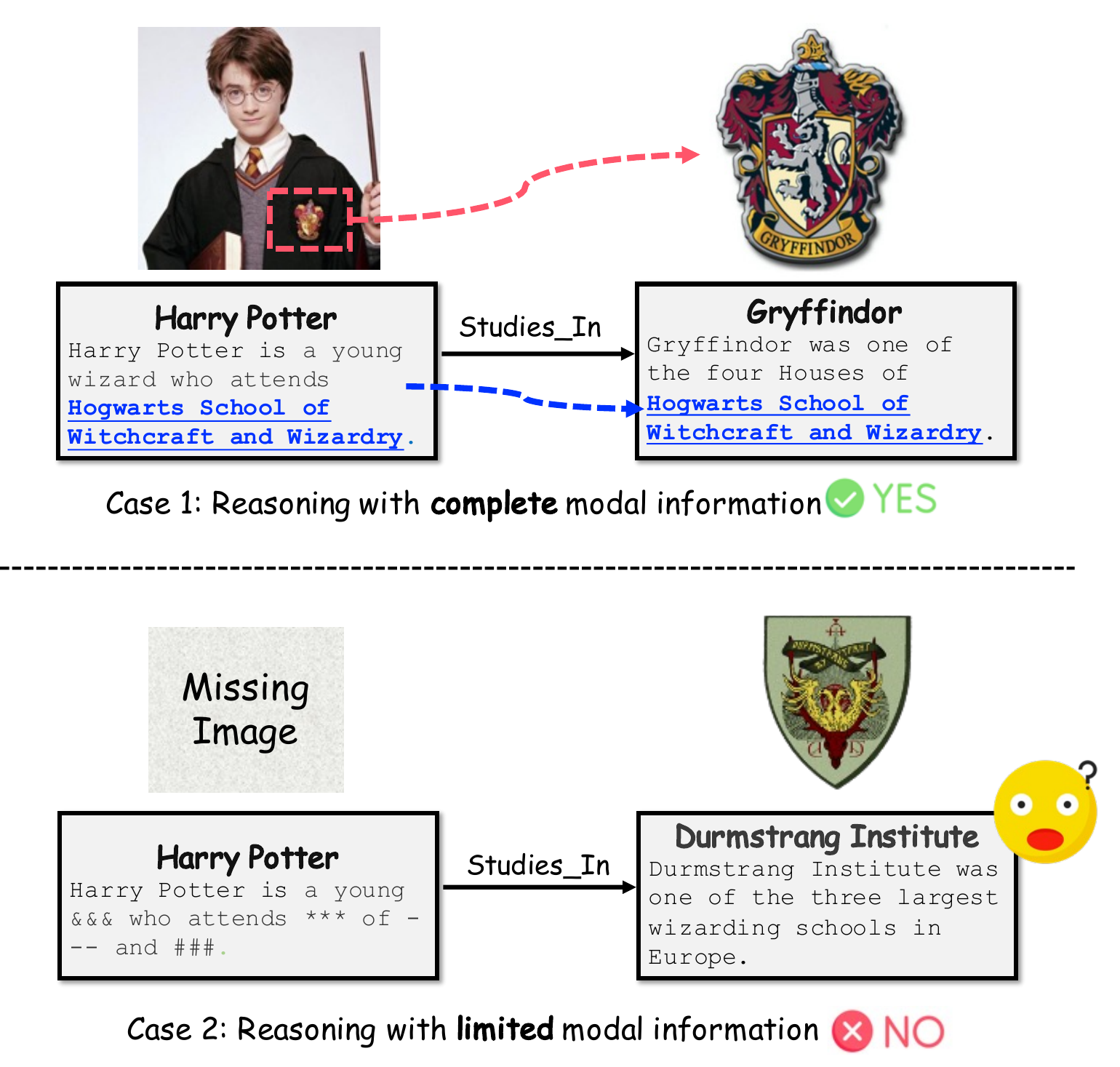}
    \caption{A simple example to show that knowledge graph reasoning with limited modal information may lead to wrong prediction.}
    \label{image::case}
    \vspace{-16pt}
\end{figure}

\par However, existing methods neglect the \textbf{imbalance of modality information} among entities, which can be observed from two perspectives. 
Firstly, as for knowledge graph reasoning, different modal information plays distinctive roles and should be adaptively considered. However, the modal fusion problem is inadequately addressed by existing methods, as the modal information is often inflexibly incorporated into the representation space of structural information uniformly. 
Secondly, the efficacious features in the images and textual descriptions are usually limited and challenging to extract. In practical scenarios, the KGs constructed from multiple heterogeneous data sources are even modality-missing, further constraining the utilization of modality information in MMKGC.  In practice, the KGs constructed with multiple heterogeneous data sources are even modality-missing \cite{DBLP:conf/mm/RSME}, which further limits the usage modality information in MMKGC. Figure \ref{image::case} reveals that limited or missing multi-modal information significantly hampers modal performance. To achieve better performance for MMKGC, it is crucial to effectively utilize the essential information as well as get higher quality multi-modal information. This need can be summarised as unleashing the power of imbalanced modality information for MMKGC.

\par To address this problem, we propose a novel MMKGC framework with \textbf{Ada}ptive \textbf{M}ulti-modal \textbf{F}usion and \textbf{M}odality \textbf{A}dversarial
\textbf{T}raining (AdaMF-MAT) to augment and effectively utilize the multi-modal information. We introduce AdaMF, an adaptive multi-modal fusion (AdaMF) module that selectively extracts essential multi-modal features from entities to generate representative joint embeddings. We further propose a modality-adversarial training (MAT) strategy to generate synthetic multi-modal embeddings and construct adversarial examples, aiming to enhance the limited multi-modal information during training. While existing NS methods design complex strategies to sample in the given KG, our method directly creates synthetic samples with semantic-rich multi-modal information to enhance the multi-modal embedding learning.
The two modules are designed to synergistically interact and mutually complement each other to unleash the power of imbalanced modality information. Our approach is a co-design of the MMKGC model and training strategy, which can bring outperforming empirical results against 19 existing unimodal KGC, multi-moda KGC, and NS methods. Our contribution is three-fold:
\begin{itemize}
    \item We propose adaptive multi-modal fusion (AdaMF) for MMKGC, to fuse the imbalanced modality information of three modalities (structural, visual, and textual) and produce representative joint embeddings.
    \vspace{-5pt}
    \item We propose a modality adversarial training strategy (MAT) to utilize imbalanced modality information. MAT aims to generate adversarial examples with synthetic multi-modal embeddings too enhance the MMKGC training.
    \vspace{-5pt}
    \item We conduct comprehensive experiments with further explorations on three public benchmarks to evaluate the performance of AdaMF-MAT. The empirical results demonstrate that AdaMF-MAT can outperform 19 recent baselines and achieve new SOTA MMKGC results. Moreover, we illustrate the suitability of AdaMF-MAT for the modality-missing scenarios.
\end{itemize}

\section{Related Works}
\subsection{Multi-modal Knowledge Graph Completion}
Existing multi-modal knowledge graph completion (MMKGC) methods \cite{Liangke_Survey, DBLP:journals/corr/MMKG-survey} are mainly concerned with extending the general knowledge graph embedding (KGE) methods \cite{DBLP:conf/nips/TransE, DBLP:journals/corr/DistMult} to the multi-modal scenario. In MMKGC, visual and textual information are delicately considered as well as the structural information of the entities to make better predictions.

\par A mainstream design approach of MMKGC methods is to enhance the entity representations with their multi-modal information.
IKRL \cite{DBLP:conf/ijcai/IKRL} first introduces the visual information of entities into the TransE \cite{DBLP:conf/nips/TransE} score function. TBKGC \cite{DBLP:conf/starsem/TBKGC} later extends IKRL with both visual and textual information. TransAE \cite{DBLP:conf/ijcnn/TransAE} employs the auto-encoder framework to fuse the multi-modal information with the entity embeddings. RSME \cite{DBLP:conf/mm/RSME} designs several gates to select and keep the useful visual information of entities. VBKGC \cite{VBKGC} uses VisualBERT to extract deep fused multi-modal information to achieve better modality fusion. OTKGE \cite{OTKGE} achieve multi-modal fusion via optimal transport. MoSE \cite{DBLP:conf/emnlp/MOSE} considered different modal information in a split-and-ensemble manner.

\par Besides, some methods propose enhanced negative sampling \cite{DBLP:conf/nips/TransE} methods to improve the MMKGC performance. MMRNS \cite{DBLP:conf/mm/MMRNS} utilizes the modal information to enhance the negative sampling to train better KGE models. MANS \cite{DBLP:journals/corr/MANS} proposes a modality-aware negative sampling method to align the structural and multi-modal information.
\subsection{Adversarial Training in Knowledge Graph Completion}
Adversarial training (AT) \cite{DBLP:conf/iclr/AML, DBLP:conf/nips/GAN, DBLP:journals/corr/GoodfellowSS14,DBLP:conf/nips/AT3} is a powerful technology widely used in many machine learning fields \cite{DBLP:journals/corr/BAGAN, DBLP:conf/iclr/ATNLP}. AT aims to train a pair of generator and discriminator with a minimax game adversarially, which can enhance both the generator and the discriminator. 
\par In the knowledge graph community, a few methods \cite{DBLP:conf/naacl/KBGAN, DBLP:conf/aaai/IGAN, DBLP:conf/ijcai/PUDA} employ AT to generate harder negative samples \cite{DBLP:conf/nips/TransE} by reinforcement learning to enhance the embedding training process and improve the prediction results. But these methods are designed for unimodal KGC.

\section{Methodology}
\begin{figure*}[]
    \centering
    \includegraphics[width=\linewidth]{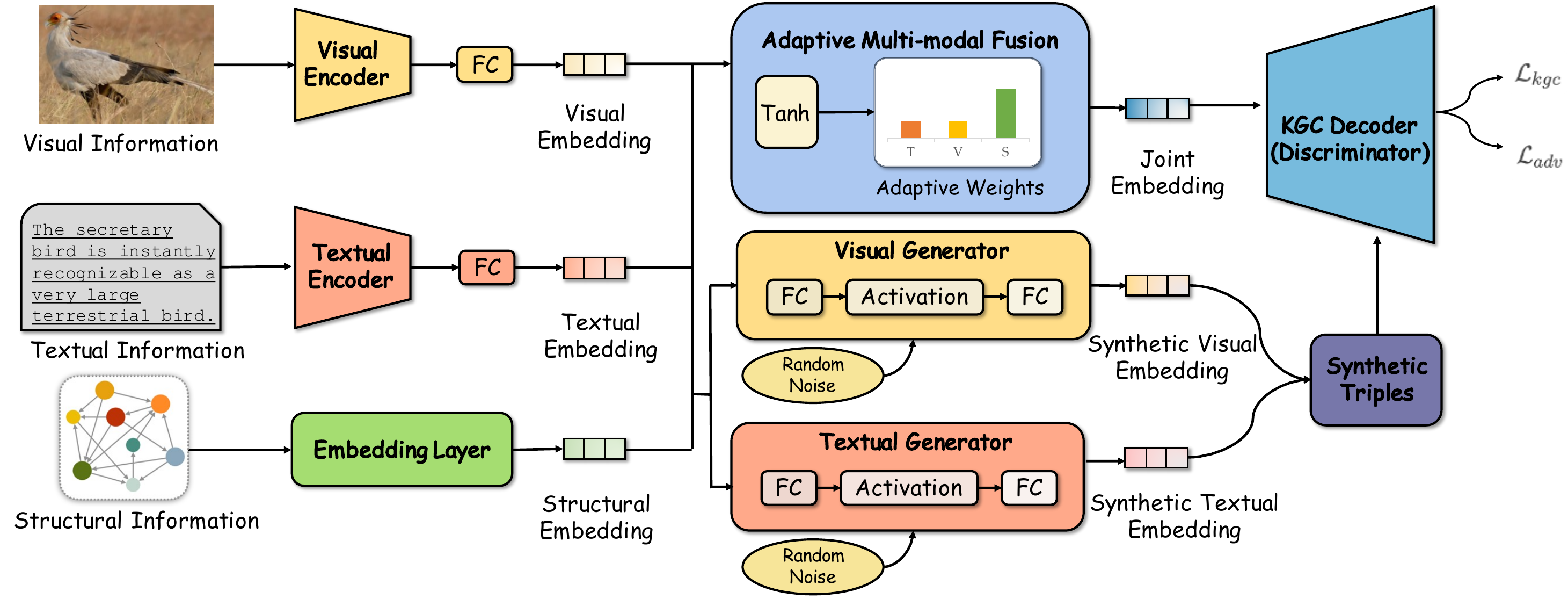}
    \caption{Overview of our method AdaMF-MAT. The feature encoders are designed to encode different modal features (visual/textual/structural) respectively. Each FC represents a fully-connected projection layer. The adaptive multi-modal fusion module is designed to get the fused joint embedding adaptively. The modality adversarial training module employs generators to generate synthetic multi-modal embeddings to construct adversarial examples. The KGC decoder serves as the discriminator and will be enhanced by these adversarial examples during training.}
    \label{image::model}
\end{figure*}

In this section, we will present our MMKGC framework AdaMF-MAT. An overview of AdaMF-MAT is shown in Figure \ref{image::model}. We will introduce the main components in the following paragraphs, which are feature encoders, adaptive multi-modal fusion, and modality adversarial training respectively.
\subsection{Preliminaries}
A knowledge graph (KG) can be denoted as $\mathcal{G}=(\mathcal{E}, \mathcal{R}, \mathcal{T})$, where $\mathcal{E}, \mathcal{R}$ are the entity set, relation set. $\mathcal{T}=\{(h, r, t)| h, t\in\mathcal{E}, r\in\mathcal{R}\}$ is the triple set. For multi-modal KGs (MMKGs), each entity $e\in \mathcal{E}$ has an image set $\mathcal{V}_e$ and a corresponding textual description.
\par An MMKGC model embeds the entities and relations into a continuous vector space. For each entity $e_i \in \mathcal{E}$, we define its structural/visual/textual embedding as $\bm{e}_{s},\bm{e}_{v},\bm{e}_{t}$ respectively to represent its different modal features. For each relation $r\in\mathcal{R}$, we denote its structural embedding as $\bm{r}$. Besides, an MMKGC model can measure the plausibility of each triple $(h, r, t)\in\mathcal{T}$ with a score function $\mathcal{F}$, which can be calculated by the defined embeddings. In the inference stage of the KGC task, for a given query $(h, r, ?)$ or $(?, r,t)$, the model ranks the corresponding triple scores of each candidate entity and make the prediction.

\subsection{Multi-modal Feature Encoding}
We first introduce the encoding process to extract the multi-modal features from the original images and textual descriptions of entities. It is a necessary step for all MMKGC methods.
\par As for the visual modal, we first apply a pre-trained visual encoder (PVE) \cite{DBLP:journals/corr/VGG,DBLP:conf/iclr/BEiT} 
to extract the visual feature $\bm{f}_v$ for each entity $e$ with mean-pooling and project it to the embedding space to get the visual embedding, which can be denoted as:

\begin{equation}
    \bm{f}_v=\frac{1}{|\mathcal{V}_e|}\sum_{img_i\in\mathcal{V}_e}\mathrm{PVE}(img_i)
\end{equation}
\begin{equation}
\bm{e}_v=\mathbf{W}_v\cdot \bm{f}_v+\mathbf{b}_v
\end{equation}
where the $\mathbf{W}_v,\mathbf{b}_v$ is the parameters of the visual project layer. 
\par For the textual embedding, we similarly extract the textual feature of each entity with its description and the pre-trained textual encoder (PTE) \cite{DBLP:conf/emnlp/SBERT}. The special [CLS] token \cite{DBLP:conf/naacl/BERT} is applied to capture the sentence-level textual feature $\bm{f}_t$, which is a common technology. Then we project the textual feature $\bm{f}_t$ to the embedding space with another project layer parameterized by $\mathbf{W}_t,\mathbf{b}_t$ to get the textual embedding $\bm{e}_t$, which can be denoted as:
\begin{equation}
    \bm{e}_t=\mathbf{W}_t\cdot \bm{f}_t+\mathbf{b}_t
\end{equation}
\par We maintain consistency with previous methods \cite{DBLP:conf/starsem/TBKGC,DBLP:conf/mm/MMRNS} and employ the same PVE and PTE of these methods for a fair comparison. Besides, the project layers are designed to project the multi-modal features from different representation spaces into the same space of the structural embeddings.

\subsection{Adaptive Multi-modal Fusion}
In MMKGC, the information of three modalities $m\in \mathcal{M}=\{s, v, t\}$ should be carefully considered to measure the triple plausibility. As existing methods \cite{DBLP:conf/ijcai/IKRL, DBLP:conf/starsem/TBKGC} usually treat them separately or simply put them together, we propose an adaptive multi-modal fusion (AdaMF for short) mechanism to achieve adaptive modality fusion. For each entity $e$ and its embeddings $\bm{e}_m$ where $m\in\mathcal{M}$, AdaMF works in the following way:
\begin{equation}
\alpha_m=\frac{\exp(\bm{w}_m\oplus\mathrm{tanh}(\bm{e}_m))}{\sum_{n\in\mathcal{M}}\exp(\bm{w}_n\oplus\mathrm{tanh}(\bm{e}_n))}
\end{equation}
\begin{equation}
\bm{e}_{joint}=\sum_{m\in\mathcal M}\alpha_{m}\bm{e}_m
\end{equation}
where $\alpha_m$ is the adaptive weight, $\oplus$ is the point-wise product operator and $\bm{w}_m$ is the learnable vector of the modality $m$. $\bm{e}_{joint}$ is the joint embedding of entity $e$. Through such a design of AdaMF, 
we can learn different modality weights for different entities, thus achieve adaptive modal information fusion to produce representative joint embeddings.

\par Besides, we apply the RotatE model \cite{DBLP:conf/iclr/RotatE} as our score function to measure the triple plausibility. It can be denoted as:
\begin{equation}
    \mathcal{F}(h, r,t)=||\bm{h}_{joint}\circ \bm{r}-\bm{t}_{joint}||
\end{equation}
where $\circ$ is the rotation operation in the complex space. The joint embeddings of the head and tail entity are employed in the score function. 

\par Aiming to assign the positve triples with higher scores, we train the embeddings with a sigmoid loss function \cite{DBLP:conf/iclr/RotatE}. The loss function can be denoted as:
\begin{equation}
    \begin{aligned}
    \mathcal{L}_{kgc}&=\frac{1}{|\mathcal{T}|}\sum_{(h, r, t)\in \mathcal{T}}\Big(-\log\sigma(\gamma-\mathcal{F}(h, r, t))\\
    &-\sum_{i=1}^{K}p_i\log\sigma(\mathcal{F}(h_i',r_i',t_i')-\gamma)\Big)
    \end{aligned}
\end{equation}
where $\sigma$ is the sigmoid function, $\gamma$ is the margin and $p_i$ is the self-adversarial weight for each negative triple $(h_i',r_i',t_i')$ generated by negative sampling \cite{DBLP:conf/nips/TransE}. The self-adversarial weight $p_i$ can be denoted as:
\begin{equation}
    p_i=\frac{\exp(\beta\mathcal{F}(h_i',r_i',t_i'))}{\sum_{j=1}^K \exp(\beta\mathcal{F}(h_j',r_j',t_j'))}
\end{equation}
where $\beta$ is the temperature parameter. A main training objective of our model is to minimize $\mathcal{L}_{kgc}$.

\subsection{Modality Adversarial Training}
Previously, we noted that the multi-modal embeddings of entities contain limited essential information and face the imbalance problem. Though AdaMF can adaptively select and fuse the multi-modal embeddings into a joint one, it can not augment the existing multi-modal embeddings to provide more semantic-rich information. 
\par Therefore, inspired by adversarial training (AT) \cite{DBLP:journals/corr/BAGAN}, we design a modality adversarial training (MAT) mechanism for MMKGC to enhance the imbalanced multi-modal information. MAT employs a generator $\mathbf{G}$ that produces adversarial samples, and a discriminator $\mathbf{D}$ that measures their plausibility, which follows the paradigm of general AT. Below we describe the design of the generator $\mathbf{G}$ and the discriminator $\mathbf{D}$.
\subsubsection{Generator}
\par Further, we design a modality adversarial generator (MAG) that assumes the role of $\mathbf{G}$ by generating synthetic multi-modal embeddings conditioned on the structural embedding $\bm{e}_{s}$, which aims to construct adversarial samples. The MAG is a two-layer feed-forward network which can be denoted as:
\begin{equation}
    \mathbf{G}_m(\bm{e}_s, z)=\mathbf{W}_2 \cdot\delta(\mathbf{W}_1\cdot[\bm{e}_s; z]+\mathbf{b}_1)+\mathbf{b}_2
\end{equation}
where $\mathbf{W}_1,\mathbf{W}_2,\mathbf{b}_1,\mathbf{b}_2$ are the parameters of the two feed-forward layers and $\delta$ is the LeakyReLU \cite{LeakyReLU} activation function. $m\in\{v, t\}$ is the visual/textual modal and $z\sim \mathcal{N}(\boldsymbol{0}, \boldsymbol{1})$ is the random noise.
\begin{algorithm}[t]
\SetKwInOut{Input}{Input}\SetKwInOut{Output}{Output}
    \caption{Pseudo-code for training AdaMF-MAT}\label{algorithm}
\KwIn{A batch of training triple $\mathcal{B}$ sampled from $\mathcal{T}$, the multi-modal information of the entities', the AdaMF model $\mathbf{D}$, the generator $\mathbf{G}$.}
\KwOut{The MMKGC model $\mathbf{D}$ trained with MAT.}
\For{each triple $(h, r, t)\in\mathcal{B}$}{// Training $\mathbf{D}$\\Get the joint embeddings $\bm{h}_{joint},\bm{t}_{joint}$.\\Calculate the triple score $\mathcal{F}(h, r, t)$ and the nagative triple scores $\mathcal{F}(h_i', r_i', t_i')$.\\ Calculate the kgc loss $\mathcal{L}_{kgc}$. \\Generate the adversarial example set $\mathcal{S}$ with $\mathbf{G}$. \\ Calculate the adversarial loss $\mathcal{L}_{adv}$.\\ Calculate the overall loss $\mathcal{L}_{kgc}+\lambda\mathcal{L}_{adv}$.
\\ Back propagation and optimize $\mathbf{D}$.\\ // Training $\mathbf{G}$.\\Get the joint embeddings $\bm{h}_{joint},\bm{t}_{joint}$.\\Generate the adversarial example set $\mathcal{S}$ with $\mathbf{G}$. \\ Calculate the adversarial loss $\mathcal{L}_{adv}$. \\ Back propagation and optimize $\mathbf{G}$.}
\end{algorithm}
\par For each triple $(h, r, t)$, we generate synthetic multi-modal embeddings for the head and tail entities and construct the synthetic entities. The synthetic head entity $h^*$  has three embeddings $\bm{h}_s,\bm{h}_v^*,\bm{h}_t^*$ where $\bm{h}_v^*,\bm{h}_t^*$ are generated by the MAG $\mathbf{G}_v$ and $\mathbf{G}_t$. 
It is the same for the tail entity.

\par Further, we can construct three synthetic triples with the two synthetic entities, which are $\mathcal{S}(h, r, t)=\{(h, r, t^*), (h^*, r, t), (h^*, r, t^*)\}$ for a triple $(h, r, t)$. Here, $\mathcal{S}$ contains a group of synthetic triples for $(h, r, t)$, which consists of 3 adversarial samples. In practice, we generate $L$ groups of synthetic triples for each $(h, r, t)$, and $\mathcal{S}$ will contain $3L$ synthetic triples. Since the generator contains random noise, these synthetic triples are different from each other.
\subsubsection{Discriminator}
\par With the generated adversarial examples with synthetic multi-modal embeddings, $\mathbf{D}$ aims to distinguish the positive (real) triples from the synthetic ones while $\mathbf{G}$ aims to generate realistic triples and deceive the discriminator $\mathbf{D}$. In MMKGC, we let the score function $\mathcal{F}$ serve as the $\mathbf{D}$ in the AT setting because $\mathcal{F}$ is the triple plausibility discriminator and the target to be adversarially enhanced. Accordingly, the loss of the adversarial training can be designed as:
\begin{equation}
    \begin{aligned}
        \mathcal{L}_{adv}&=\frac{1}{|\mathcal{T}|}\sum_{(h, r, t)\in \mathcal{T}}\Big(-\log\sigma(\gamma-\mathcal{F}(h, r, t))\\
        &-\frac{1}{|\mathcal{S}|}\sum_{(h^*,r^*,t^*)\atop\in\mathcal{S}(h, r, t)}\log\sigma(\mathcal{F}(h^*,r^*,t^*)-\gamma)\Big)
    \end{aligned}
\end{equation}

\par We keep the format of the loss function similar to $\mathcal{L}_{kgc}$ and contrast the positive triples and the adversarial examples, as such a format is more suitable for MMKGC. From another perspective, MAT can be regarded as an enhanced negative sampling method for MMKGC, as we generate high-quality negative examples in an adversarial manner. The main difference between MAT and existing negative sampling methods like MMRNS\cite{DBLP:conf/mm/MMRNS}, and KBGAN \cite{DBLP:conf/naacl/KBGAN} is that MAT generates new hard negative samples by adversarial training instead of sampling from the existing KGs.  Such a design of MAT can directly augment the multi-modal embedding learning and enhance the performance of $\mathbf{D}$ without the need for reinforcement learning applied in KBGAN \cite{DBLP:conf/naacl/KBGAN}. Besides, MMRNS \cite{DBLP:conf/mm/MMRNS} and KBGAN \cite{DBLP:conf/naacl/KBGAN} just employ the unimodal score functions but our approach introduces the multi-modal information with adaptive multi-modal fusion, which is a co-design of the MMKGC model and training strategy.

\subsubsection{Training Ojective}
During training, we iteratively train the $\mathbf{D}$ and $\mathbf{G}$ in an adversarial manner while keeping to optimize the general objective $\mathcal{L}_{kgc}$ for MMKGC. The overall training objective can be represented as:
\begin{equation}
\min_{\mathbf{D}}\mathcal{L}_{kgc}+\min_{\mathbf{D}}\max_{\mathbf{G}}\lambda \mathcal{L}_{adv}
\end{equation}
where $\lambda$ is the coefficient of the adversarial loss. 
The pseudo-code of mini-batch training for AdaMF-MAT is shown in Algorithm \ref{algorithm}.

\par During training, the AdaMF model $\mathbf{D}$ and the generator $\mathbf{G}$ will be optimized iteratively in an adversarial setting.

\begin{table}[]
\caption{Statistical information of the benchmarks.}
\label{table::dataset}
\centering
\resizebox{\columnwidth}{!}{
\begin{tabular}{c|ccccc}
\toprule
Dataset   & $|\mathcal{E}|$    & $|\mathcal{R}|$   & \#Train & \#Valid & \#Test \\ \midrule
DB15K & 12842 & 279 & 79222  & 9902   & 9904  \\
MKG-W    & 15000 & 169  & 34196   & 4276    & 4274   \\
MKG-Y    & 15000 & 28  & 21310   &  2665   & 2663   \\
\bottomrule
\end{tabular}}
\end{table}
\begin{table*}
\resizebox{\textwidth}{!}{
\begin{tabular}{cc|cccc|cccc|cccc}
\toprule
\multicolumn{2}{c|}{\multirow{2}{*}{Model}} & \multicolumn{4}{c|}{DB15K} & \multicolumn{4}{c|}{MKG-W} & \multicolumn{4}{c}{MKG-Y} \\
\multicolumn{2}{c|}{} & MRR & Hit@1 & Hit@3 & Hit@10& MRR & Hit@1 & Hit@3 & Hit@10& MRR & Hit@1 & Hit@3 & Hit@10\\
\midrule
\multicolumn{1}{c|}{\multirow{6}{*}{\begin{tabular}[c]{@{}c@{}}Unimodal\\ KGC\end{tabular}}}& TransE& 24.86 & 12.78 & 31.48 & 47.07 & 29.19 & 21.06 & 33.20& 44.23 & 30.73 & 23.45 & 35.18 & 43.37 \\
\multicolumn{1}{c|}{} & TransD& 21.52 & 8.34& 29.93 & 44.24 & 25.56 & 15.88 & 32.99 & 40.18 & 26.39 & 17.01 & 33.60 & 40.31 \\
\multicolumn{1}{c|}{} & DistMult& 23.03 & 14.78 & 26.28 & 39.59 & 20.99 & 15.93 & 22.28 & 30.86 & 25.04 & 19.33 & 27.80 & 35.95 \\
\multicolumn{1}{c|}{} & ComplEx & 27.48 & 18.37 & 31.57 & 45.37 & 24.93 & 19.09 & 26.69 & 36.73 & 28.71 & 22.26 & 32.12 & 40.93 \\
\multicolumn{1}{c|}{} & RotatE& 29.28 & 17.87 & 36.12 & 49.66 & 33.67 & 26.80 & 36.68 & 46.73 & 34.95 & 29.10 & 38.35 & 45.30 \\
\multicolumn{1}{c|}{} & PairRE& 31.13 & 21.62 & 35.91 & 49.30 & 34.40& 28.24 & 36.71 & 46.04 & 32.01 & 25.53 & 35.84 & 43.89 \\
\multicolumn{1}{c|}{} & GC-OTE& 31.85 & 22.11 & 36.52 & 51.18 & 33.92 & 26.55 & 35.96 & 46.05 & 32.95 & 26.77 & 36.44 & 44.08 \\
\midrule
\multicolumn{1}{c|}{\multirow{7}{*}{\begin{tabular}[c]{@{}c@{}}Multi-modal\\ KGC\end{tabular}}}& IKRL & 26.82 & 14.09 & 34.93 & 49.09 & 32.36 & 26.11 & 34.75 & 44.07 & 33.22 & 30.37 & 34.28 & 38.26 \\
\multicolumn{1}{c|}{} & TBKGC& 28.40 & 15.61 & 37.03 & 49.86 & 31.48 & 25.31 & 33.98 & 43.24 & 33.99 & 30.47 & 35.27 & 40.07 \\
\multicolumn{1}{c|}{} & TransAE & 28.09 & 21.25 & 31.17 & 41.17 & 30.00 & 21.23 & 34.91 & 44.72 & 28.10 & 25.31 & 29.10 & 33.03 \\
\multicolumn{1}{c|}{} & MMKRL& 26.81 & 13.85 & 35.07 & 49.39 & 30.10 & 22.16 & 34.09 & 44.69 & 36.81 & 31.66 & 39.79 & 45.31 \\
\multicolumn{1}{c|}{} & RSME & 29.76 & {\ul 24.15}& 32.12 & 40.29 & 29.23 & 23.36 & 31.97 & 40.43 & 34.44 & 31.78 & 36.07 & 39.09 \\
\multicolumn{1}{c|}{} & VBKGC& 30.61 & 19.75 & 37.18 & 49.44 & 30.61 & 24.91 & 33.01 & 40.88 & 37.04 & {\ul 33.76} & 38.75 & 42.30 \\
\multicolumn{1}{c|}{} & OTKGE& 23.86 & 18.45 & 25.89 & 34.23 & 34.36 & {\ul28.85} & 36.25 & 44.88 & 35.51 & 31.97 & 37.18 & 41.38 \\
\midrule
\multicolumn{1}{c|}{\multirow{6}{*}{\begin{tabular}[c]{@{}c@{}}Negative\\ Sampling\end{tabular}}} 
& KBGAN(TransE) & 25.73 & 9.91 & 36.95& 51.93& 29.47 & 22.21 & 34.87 & 40.64 & 29.71 & 22.81 & 34.88 & 40.21 \\
\multicolumn{1}{c|}{}& KBGAN(TransD) & 23.74 & 9.34 & 33.51 & 47.94 & 29.67 & 22.38 & 35.24 & 40.80 & 28.73 & 20.99 & 34.64 & 40.76 \\
\multicolumn{1}{c|}{}& MANS & 28.82 & 16.87 & 36.58& 49.26& 30.88 & 24.89 & 33.63 & 41.78 & 29.03 & 25.25 & 31.35 & 34.49 \\
\multicolumn{1}{c|}{}& MMRNS(RotatE) & 29.67 & 17.89 & 36.66 & 51.01 & 34.13 & 27.37 & 37.48 & 46.82 & 35.93 & 30.53 & 39.07 & 45.47 \\
\multicolumn{1}{c|}{} & MMRNS(SOTA) & {\ul 32.68}& 23.01& 37.86 & 51.01 & {\ul 35.03}& 28.59 &37.49& {\ul 47.47}& 35.93 & 30.53 & 39.07 & 45.47 \\
\midrule
\multicolumn{1}{c|}{\multirow{2}{*}{Ours}}& AdaMF& 32.51 & 21.31 & {\ul 39.67}& {\ul 51.68}& 34.27	&27.21	& {\ul 37.86} &	47.21 & {\ul 38.06}& 33.49& {\ul 40.44}& {\ul 45.48}\\
\multicolumn{1}{c|}{} & AdaMF-MAT & \textbf{35.14} & \textbf{25.30} & \textbf{41.11} & \textbf{52.92} & \textbf{35.85} & 	\textbf{29.04}	& \textbf{39.01} &	\textbf{48.42} & \textbf{38.57} & \textbf{34.34} & \textbf{40.59} & \textbf{45.76}\\
\bottomrule
\end{tabular}}

\caption{Link prediction performance on DB15K, MKG-W, and MKG-Y. The best results are marked \textbf{bold} and the second-best results are {\ul underlined} in each column. The results of unimodal KGE and MMRNS are from \cite{DBLP:conf/mm/MMRNS} while other results are based on our reproduction.
MMRNS(SOTA) is the state-of-the-art results of MMRNS among different score functions.}
\label{table::main}
\end{table*}
\section{Experiments}
In this section, we evaluate our method with the link prediction task, which is a mainstream task in KGC.
We first introduce the experiment settings and then present our experimental results. We mainly explore the following four research questions (RQ) about AdaMF-MAT:
\begin{itemize}
    \item \textbf{RQ1}: How does the performance of AdaMF-MAT compare to the existing MMKGC methods in the link prediction task?
    \item \textbf{RQ2}: How does the performance of AdaMF-MAT in the modality-missing scenario?
    \item \textbf{RQ3}: How much do the design of AdaMF and MAT contribute to the performance?
    \item \textbf{RQ4}: can we find some intuitive cases to explain the performance of our method?
\end{itemize}
\vspace{-8pt}
\subsection{Experiment Settings}
\subsubsection{Datasets}
The experiments are conducted on three public benchmarks. They are DB15K, MKG-W, and MKG-Y. DB15K \citetlanguageresource{DBLP:conf/esws/MMKG} is constructed from DBPedia \citetlanguageresource{DBLP:journals/semweb/DBPedia} with images crawled from a search engine. MKG-W and MKG-Y are proposed by \cite{DBLP:conf/mm/MMRNS}, which are the subsets of Wikidata \citetlanguageresource{DBLP:journals/cacm/wikidata} and YAGO \citetlanguageresource{DBLP:conf/www/yago} knowledge bases respectively. The statistical information about the three datasets is shown in Table \ref{table::dataset}. We reuse the multi-modal features captured with pre-trained models \cite{DBLP:journals/corr/VGG,DBLP:conf/naacl/BERT} released by the original papers.

\subsubsection{Task and Protocols}

To evaluate our method, we conduct link prediction \cite{DBLP:conf/nips/TransE} task on the three datasets. Link prediction is a significant task of knowledge graph completion, which aims to predict the missing entity of a given query $(h,r,?)$ or $(?, r, t)$. The two parts of the link prediction task is called head prediction and tail prediction respectively.
\par Following the existing works, we use rank-based metrics \cite{DBLP:conf/iclr/RotatE} like mean reciprocal rank (MRR) and Hit@K(K=1, 3, 10) to evaluate the results. 
MRR and Hit@K can be calculated as :
\begin{equation}
   \mathbf{MRR}=\frac{1}{|\mathcal{T}_{test}|}\sum_{i=1}^{|\mathcal{T}_{test}|}(\frac{1}{r_{h,i}}+\frac{1}{r_{t,i}})
\end{equation}
\begin{equation}
   \mathbf{Hit@K}=\frac{1}{|\mathcal{T}_{test}|}\sum_{i=1}^{|\mathcal{T}_{test}|}(\mathbf{1}(r_{h,i} \leq K)+\mathbf{1}(r_{t,i} \leq K))
\end{equation}
where $r_{h,i}$ and $r_{t,i}$ are the results of head prediction and tail prediction respectively.
\par Besides, filter setting \cite{DBLP:conf/nips/TransE} is applied to all the results to avoid the influence of the training triples for fair comparisons.

\subsubsection{Baselines}
To demonstrate the effectiveness of our approach, we select several MMKGC methods as our baselines for comparisons and they can be divided into three categories:
\par {{(\rmnum{1})}} \textbf{Uni-modal KGC methods}, including TransE \cite{DBLP:conf/nips/TransE}, TransD \cite{DBLP:conf/acl/TransD},  DistMult \cite{DBLP:journals/corr/DistMult}, ComplEx \cite{DBLP:journals/jmlr/Complex}, RotatE \cite{DBLP:conf/iclr/RotatE}, PairRE \cite{DBLP:conf/acl/PairRE}, and GC-OTE \cite{DBLP:conf/acl/OTE}. These methods only use the structural information of the KGs to learn structural embeddings.
\par {{(\rmnum{2})}} \textbf{Multi-modal KGC models}, including IKRL \cite{DBLP:conf/ijcai/IKRL}, TBKGC \cite{DBLP:conf/starsem/TBKGC}, TransAE \cite{DBLP:conf/ijcnn/TransAE}, MMKRL \cite{DBLP:journals/apin/MMKRL}, RSME \cite{DBLP:conf/mm/RSME}, VBKGC \cite{VBKGC}, OTKGE \cite{OTKGE}. These methods utilize both the structural information and multi-modal information in the KGs.
\par {{(\rmnum{3})}} \textbf{Negative sampling methods}, including KBGAN \cite{DBLP:conf/naacl/KBGAN},
MANS \cite{DBLP:journals/corr/MANS} and MMRNS \cite{DBLP:conf/mm/MMRNS}. KBGAN is the first work for adversarial negative sample in KGC based on reinforcement learning. MANS and MMRNS design new negative sampling strategies enhanced by the multi-modal information to generate high-quality negative samples.
    
\par Besides, we re-implement some baselines that have no officially released codes and conduct the link prediction experiments according to the detailed experimental settings in the original paper.

\subsubsection{Implemention Details}
We implement our method based on OpenKE \cite{DBLP:conf/emnlp/OpenKE} and conduct experiments on Ubuntu 22.04.1 operating system with NVIDIA GeForce 3090 GPUs. We employ VGG \cite{DBLP:journals/corr/VGG} for DB15K and BEiT \cite{DBLP:conf/iclr/BEiT} for MKG-W/MKG-Y as PVEs. As for the textual modal, we employ sentence-BERT \cite{DBLP:conf/emnlp/SBERT} following the setting in MMRNS \cite{DBLP:conf/mm/MMRNS}.
\par For the hyper-parameters, we set the embedding dimension to 200 while the negative sample number $K$ is tuned in $\{32, 64, 128\}$. The dimension of the random noise in the MAT is 64. The group number of adversarial samples $L$ is set to be 1. The training batch size is fixed to 1024. We search the margin $\gamma\in\{1, 2, 4, 8, 12\}$, the temperature $\beta \in \{0.5, 1.0, 2.0\}$ and the learning rate $\eta_d, \eta_g \in$$ \{10^{-3}, 10^{-4}, 10^{-5}\}$ for $\mathbf{D}$ and $\mathbf{G}$. The coefficient $\lambda$ of the adversarial loss is searched in $\{0.1, 0.01, 0.001\}$.

\subsection{Main Results (RQ1)}
To address \textbf{RQ1}, we conduct the main link prediction experiments and present the results in Table \ref{table::main}. Our method AdaMF-MAT can outperform the existing 19 MMKGC baselines, with an average improvement of 6.0\% in MRR, 8.0\% in Hit@1, 5.5\% in Hit@3, and 2.1\% in Hit@10. These results indicate that AdaMF-MAT shows more significant improvements, particularly on strict metrics like Hit@1, which means that AdaMF-MAT can make more exact predictions.
\par Compared with MMRNS(RotatE) which also employs RotatE \cite{DBLP:conf/iclr/RotatE} as the score function, our AdaMF (w/o MAT) can achieve better results on all metrics. Such a result demonstrates that AdaMF can do better than the existing methods to utilize the multi-modal information in the MMKGs. Additionally, we can observe that MAT can bring full range improvement of metrics by generating high-quality synthetic triples for positive-negative contrast during training, compared with other negative sampling methods like MANS and MMRNS$^2$. Some old adversarial methods like KBGAN may even lead to a decrease in performance, which means MAT is more stable.

\subsection{Modality-missing Results (RQ2)}
\begin{figure}[]
    \centering
    \includegraphics[width=\linewidth]{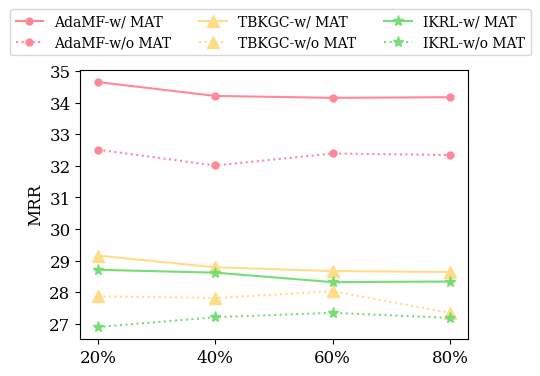}
    \caption{Link prediction results on modality-missing DB15K dataset. The x-axis represents the modality-missing ratios. We report the MRR results of three various MMKGC models (AdaMF, TBKGC, IKRL). The missing modal information is randomly initialized first, as commonly done in existing methods \cite{DBLP:conf/starsem/TBKGC}. w/ MAT and w/o MAT indicate the MMKGC model trained with MAT and without MAT respectively.}
    \label{image::missing}
\end{figure}
\par To answer \textbf{RQ2}, we conduct modality-missing link prediction experiments on DB15K datasets. In the experiments, the multi-modal information is dropped with a given modality-missing ratio. We first complete the missing information with random initialization as commonly done in existing methods \cite{DBLP:conf/starsem/TBKGC} and make comparisons of the link prediction performance among different MMKGC models (AdaMF, TBKGC, IKRL) and training strategies (w/ MAT and w/o MAT). The MRR results are shown in Figure \ref{image::missing}. We can draw the following three conclusions from the  results:
\par {{(\rmnum{1})}} AdaMF can achieve better link prediction performance in the modality-missing KGs compared with other existing MMKGC baselines like IKRL and TBKGC. This suggests that the adaptive multi-modal fusion performs better than the existing modal fusion methods when the entity modal information is missing.
\par {{(\rmnum{2})}} MAT can also enhance the MMKGC model to learn better multi-modal embeddings by adversarial training and can be applied to different MMKGC models as a general framework. The results indicate that MAT can improve the performance of all three MMKGC models compared with randomly initializing the missing modal information, which is a vanilla approach in the modality-missing scenario wide used by existing methods \cite{DBLP:conf/starsem/TBKGC,DBLP:conf/mm/RSME}.
\par {{(\rmnum{3})}} For the modality-missing KGs, the link prediction results of the MMKGC model using random initialization are not stable. Figure \ref{image::missing} shows that sometimes the experimental results increase even when the missing ratio decreases. We think this is because the various MMKGC models can only make modal fusion but can not augment the limited modal information. When the missing modal information exceeds a certain range, such modal fusion may fail and the results might be unstable. However, MAT resolves such a problem and enables MMKGC models to achieve better performance and
make full use of the modal information, as the performance is negatively correlated with the missing ratio. When the modal information is more sufficient, the model's capability will be further enhanced by utilizing more modal information.
\par 
In summary, the experimental results prove that our AdaMF-MAT is a better choice for MMKGC in the modality-missing scenario.

\subsection{Ablation Study (RQ3)}
\begin{table}[h]
\resizebox{\columnwidth}{!}{
\begin{tabular}{cc|cccc}
\toprule
\multicolumn{2}{c|}{Model}                                    & MRR   & Hit@1 & Hit@3 & Hit@10 \\
\midrule
\multicolumn{2}{c|}{AdaMF-MAT}                               & 35.14 & 25.30 & 41.11 & 52.92  \\
\midrule
\multicolumn{1}{c|}{\multirow{5}{*}{\begin{tabular}[c]{@{}c@{}}AdaMF\\(w/o MAT)\end{tabular}}} & S+V+T(Adaptive) & 33.19 & 23.08 & 40.34 & 52.47  \\
\multicolumn{1}{c|}{}                       & S+V+T(Mean)     & 32.57 & 21.45 & 39.71 & 51.68  \\
\multicolumn{1}{c|}{}                       & S+V(w/o T)      & 32.34 & 21.84 & 38.90  & 50.76  \\
\multicolumn{1}{c|}{}                       & S+T(w/o V)      & 31.82 & 19.63 & 39.69 & 52.51  \\
\multicolumn{1}{c|}{}                       & V+T(w/o S)      & 31.01 & 18.45 & 39.38 & 52.27  \\
\midrule
\multicolumn{1}{c|}{\multirow{3}{*}{MAT}}   & w/o $(h^*,r,t)$           & 34.64 & 24.52 & 40.98 & 52.49  \\
\multicolumn{1}{c|}{}                       & w/o $(h,r,t^*)$           & 34.65 & 24.49 & 41.13 & 52.61  \\
\multicolumn{1}{c|}{}                       & w/o $(h^*,r,t^*)$         & 34.61 & 24.36 & 40.98 & 52.65 \\
\bottomrule
\end{tabular}}
\caption{The ablation study results. S/V/T represent structural/visual/textual embeddings respectively. Adaptive and mean represent the different settings of the weights of each modality in the joint embeddings.}
\label{table::abl}
\end{table}
To demonstrate in more detail the effectiveness of each module design and answer \textbf{RQ3}, we conduct the ablation study as shown in Table \ref{table::abl}. We validate the effectiveness of AdaMF by replacing the adaptive weights with mean weights for different modalities. Besides, we remove the embeddings of each modality to validate the contribution of modal information. As for  MAT, we remove each kind of adversarial example to check their effectiveness. 
\par From the ablation study results we can find that the link prediction results decrease when each module or each modal information of our method is removed, which demonstrates their effectiveness.
\par
\begin{figure}[]
    \centering
    \includegraphics[width=0.8\linewidth]{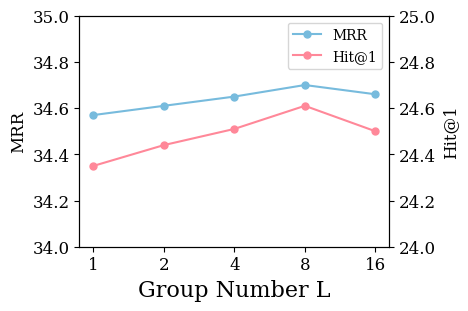}
    \caption{Parameter analysis results about the number of adversarial examples. We report the MRR and Hit@1 results.}
    \label{image::parameter}
\end{figure}

 To better understand MAT, we conduct parameter analysis on the group number $L$ of the adversarial examples. As shown in Figure \ref{image::parameter}, we can observe that the model performance can be affected to some extent by $L$, but there is not a very conspicuous change. As $L$ increases, the link prediction performance first increases a few and then decreases.

\subsection{Case Study (RQ4)}
\begin{figure}[t]
    \centering
    \includegraphics[width=0.95\linewidth]{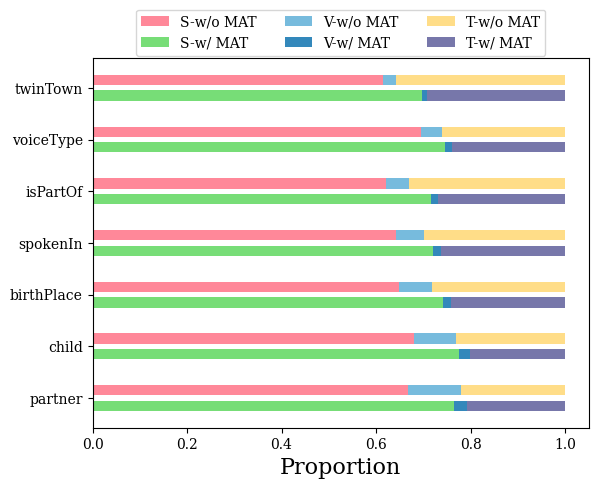}
    \caption{Adaptive weight visualization results of AdaMF for different relations. We consider the AdaMF models trained w/ and w/o MAT. We partitioned the test triples by relation and calculated the average modality weights among the entities.}
    \label{image::attention}
    \vspace{-4pt}
\end{figure}
To demonstrate the effectiveness of AdaMF-MAT in an intuitive view, we visualize the percentage of modality weights in AdaMF under several relations. As shown in Figure \ref{image::attention}, we can observe that the modality weights of entities in different relational contexts vary, and the importance of the three modalities is often $s > v > t$. This reflects that 
AdaMF can extract diverse modal features for different entities and the modality importance is also consistent with the previous research results \cite{DBLP:conf/mm/RSME,DBLP:conf/mm/MMRNS}.
\par Meanwhile, with the enhancement of MAT, the modality weights are changed compared with a vanilla AdaMF model. The overall trend is a further decrease in the weight of visual/textual embeddings especially visual modality, which indicates that MAT can augment the multi-modal embeddings and achieve better adaptive modal fusion. With MAT, the truly essential part of the multi-modal information is augmented and retained to participate in the score function.
\vspace{-8pt}
\section{Conclusion}
In this paper, we mainly discuss the problem of utilizing modal information in MMKGC and propose a novel MMKGC framework called AdaMF-MAT to address the limitations of the existing methods. Existing methods for utilizing modal features are relatively crude and treat the modal information in a one-size-fits-all manner. Meanwhile, the modality-missing problem is ignored by existing methods, which we think is also a significant problem in MMKGC. Our method AdaMF-MAT employs adaptive modal fusion to utilize the multi-modal information diversely and augment the multi-modal embeddings through modality-adversarial training. Experiments demonstrate that AdaMF-MAT can outperform all the existing baseline methods and achieve SOTA results in MMKGC tasks. 
\par We think that the utilization problem of the multi-modal information in the MMKGs is still considerable. The combination of large language models and MMKGs is a new and popular approach to MMKGC and the downstream usage of MMKGs.

\vspace{-8pt}
\section*{Limitations and Ethics}
Our work mainly proposes an adaptive fusion and adversarial training framework for MMKGC. We think the limitation of this work is that our settings for modality-missing scenarios are not close enough to real production scenarios. Besides, we need to design a more effective approach to utilize the multi-modal information of MMKGs. All experiments were conducted on publicly available datasets, with no violation of scientific ethics or invasion of privacy involved.

\section*{Acknowledgements}
This work is founded by National Natural Science Foundation of China ( NSFC62306276 / NSFCU23B2055 / NSFCU19B2027 / NSFC91846204 ), Zhejiang Provincial Natural Science Foundation of China (No. LQ23F020017), Yongjiang Talent Introduction Programme (2022A-238-G),  Fundamental Research Funds for the Central Universities (226-2023-00138).


\nocite{*}
\section*{References}
\label{sec:reference}

\bibliographystyle{lrec-coling2024-natbib}
\bibliography{lrec-coling2024-example}

\section*{Language Resource References}
\label{lr:ref}
\bibliographystylelanguageresource{lrec-coling2024-natbib}
\bibliographylanguageresource{languageresource}

\end{document}